# The Impact of Ontology on the Prediction of Cardiovascular Disease Compared to Machine Learning Algorithms

https://doi.org/10.3991/ijoe.v18i11.32647

Hakim El Massari[1(✉)], Noreddine Gherabi[1], Sajida Mhammedi[1], Hamza Ghandi[1], Mohamed Bahaj[2], Muhammad Raza Naqvi[3]
[1] National School of Applied Sciences, Sultan Moulay Slimane University, Beni-Mellal, Morocco
[2] Faculty of Sciences and Technologies, Hassan First University, Settat, Morocco
[3] National Engineering School of Tarbes INP- ENIT, University of Toulouse, Toulouse, France
h.elmassari@usms.ma

**Abstract—**Cardiovascular disease is one of the chronic diseases that is on the rise. The complications occur when cardiovascular disease is not discovered early and correctly diagnosed at the right time. Various machine learning approaches, including ontology-based Machine Learning techniques, have lately played an essential role in medical science by building an automated system that can identify heart illness. This paper compares and reviews the most prominent machine learning algorithms, as well as ontology-based Machine Learning classification. Random Forest, Logistic regression, Decision Tree, Naive Bayes, k-Nearest Neighbours, Artificial Neural Network, and Support Vector Machine were among the classification methods explored. The dataset used consists of 70000 instances and can be downloaded from the Kaggle website. The findings are assessed using performance measures generated from the confusion matrix, such as F-Measure, Accuracy, Recall, and Precision. The results showed that the ontology outperformed all the machine learning algorithms.

## 1 Introduction

Cardiovascular diseases (CVD) are a group of disorders affecting the heart and blood vessels. According to the World Health Organization (WHO), CVD is the leading cause of death in the world more people die each year from CVD than from any other cause. An estimated 17.7 million deaths are attributable to CVD, representing 31% of total global mortality. Given these statistical numbers, it is important to reveal cardiovascular disease as early as possible with the help of the trending technology based on machine learning and ontology, so that management with assistance and medicines can begin.

Machine learning (ML) is one of the most constantly evolving areas of computer science, with a wide range of applications. It is the process of obtaining usable infor-mation from a big quantity of data. Medical diagnosis, marketing, industry, and other scientific domains all make use of ML approaches. ML algorithms are well-suited for medical data analysis since they have been frequently employed in medical datasets. ML comes in several forms, including classification, regression, and clustering. Each form has a particular consequence and influence depending on the problem that we are attempting to address. We focus on classification algorithms in our work because of their high accuracy and performance in classifying a given dataset into predetermined categories and predicting future events or information from that data. In the medical field, classification algorithms are often utilized, particularly in the diagnosis of illnesses such as cardiovascular disease. Therefore, the commonly used machine learning classification [1] namely SVM, NB, DT, KNN, ANN, and LR are applied to identify patients with cardiovascular disease at an early period.

On the other hand, ontology has been one of the most widely used techniques to managing, organizing, and extracting data during the last few decades. It is a way of data representation that has been effectively utilized in a number of domains, particularly the medical domain. It is significant in computer science because of its ability to express many concepts and their relationships across fields. In reality, no single ontology is sufficient to meet today's expanding healthcare demands, and ontologies must be combined with machine learning algorithms to facilitate data integration and analysis. In previous work [2], [3], we already created and explored an ontology-based decision tree model able to predict diabetes.

In this paper, we aim to make a comparative analysis among the seven popular classification techniques and ontology-based machine learning classification based on carefully chosen parameters such as Precision, Accuracy, Recall, and F-Measure, which are derived from the confusion matrix.

The organization of the remainder of the paper is as follows: Sect. 2 represents the literature review of related classification algorithms in the field of cardiovascular prediction. Sect. 3 we present methods used in this comparative analysis and the performance metrics used to evaluate the models. In Sec. 4, we give the findings and discuss them. Finally, Sec. 5 discusses future research and conclusions.

## 2 Literature review

Lately, researchers released a substantial amount of research using machine-learning approaches to detect individuals at risk of cardiovascular disease based on symptoms. [4], [5] These techniques were shown to be impartial and beneficial. Different approaches are applied in heart disease prediction field [6], such as the Support Vector Machines (SVM), Random Forest (RF) Algorithm, Artificial Neural Networks (ANN), Particle Swarm Optimization (PSO), Genetic algorithm (GA), Naive Bayes (NB), K-Nearest Neighbor (KNN), and Decision Trees (DT) [7]–[9]. In this part, we shall discuss the most recent of them.

In this experimental analysis [10] five machine learning algorithms Random Forest, Decision Tree, K-Nearest Neighbors, Support Vector Machine, and Naive Bayes are used in the predictive analysis of early-stage of heart disease. The dataset used contains 4241 instances. High accuracy of 84.08% and 83.96% goes to the SVM and Naïve Bayes respectively.

To predict heart disease [11], the authors worked on a heart disease dataset collected from Kaggle and used k-nearest neighbor (KNN), decision tree (DT), random forests (RF), AdaboostM1 (ABM1), Logistic regression (LR), and Multilayer perceptron (MLP) algorithms. 100% of accuracy achieved by KNN, RF and DT classifiers.

This overview article [12] is a compilation of work done on the subject of Cardiovascular Disease Prediction Using Machine and Deep Learning Techniques. [13] In another study, the authors use two machine learning algorithms, SVM and ANN, to detect heart illness early, and the high-accuracy prediction findings are sent to support vector machine.

[14] The authors used NB, KNN, RF, and DT data mining classification approaches to predict heart disease. The algorithm that provides the greatest results in this model is K-nearest neighbor.

[15] In their paper, they proposed a model using Machine Learning algorithms and Relief and LASSO Feature Selection to build new hybrid classifiers based on Random Forest, AdaBoost Boosting Method, Decision Tree, K-Nearest Neighbors, and Gradient Boosting. The accuracy rate of 99.35 % goes to Random Forest, which is higher than the other algorithms.

[16] The study applied seven machine learning classification algorithms (Naive Bayes, K-Nearest Neighbors, Multi-Layer Perceptron, Logistic Regression, Decision Tree, Random Forest, and Support Vector Machine) on a dataset of cardiovascular disease. Multi-Layer Perceptron got the best accuracy result of 87.23%.

The authors [17] provide a different way of identifying key characteristics using machine learning techniques, which will improve the accuracy of cardiovascular disease prediction. The prediction model is provided with many feature combinations and numerous well-known classification approaches. We get an improved performance level with an accuracy level of 88.7 percent by combining the hybrid random forest with a linear model in the prediction model for heart disease.

In this experimental analysis [18] seven ML algorithms, HRFLM, Logistic Regression, RF, DT, K-nearest neighbor, SVM, and Linear Model are used in the predictive analysis of heart disease. High accuracy of 88.7% was attained using the Hybrid Random Forest with a Linear Model (HRFLM) technique.

In [19], the researchers used machine-learning algorithms including SMO, Multilayer Perceptron, Bayesian Network, and Random Forest. The Bayesian Network classifier performs better and achieved a 94.5 % accuracy, which is higher than the other three algorithms. In another case study based in Iraq [20], the high rate of accuracy of 94.5% goes for the Decision tree.

In [21], the authors used two datasets to predict cardiovascular with machine learning algorithms such as DT, NB, AdaBoost, SVM, RF, ANN, k-Nearest, and LR, k-NN. The results show that SVM outperforms in terms of an accuracy rate of 84%.

[22] The authors create a healthcare application to aid in the detection of cardiac disorders in patients and those experiencing symptoms. Their work is more accurate since it is based on the random forest method.

In [23], the authors compared the accuracy of the classifiers: Artificial Neural Network, Support Vector Machine, Naïve Bayes, Decision Tree, and Random Forest, Artificial Neural Network got 84.25 % compared to the other classifiers.

Furthermore, Deep Learning is used to develop complicated prediction models and has been effectively applied to a variety of difficulties in healthcare in particular. Among these, the authors of [24] provide a predictive deep learning model for heart disease prediction using RNN, GRU, and LSTM. A high level of precision of 98 percent was attained. The authors of [25] suggested a model with good accuracy, based on a two-stage deep learning LSTM neural network, to classify arrhythmias.

## 3    Methods and evaluation

In this section we present the methodologies and materials used, as well as the experimental workflow, dataset description, machine learning algorithms, ontology model, and evaluation metrics. The experimental workflow of this comparative analysis is illustrated in Figure 1.

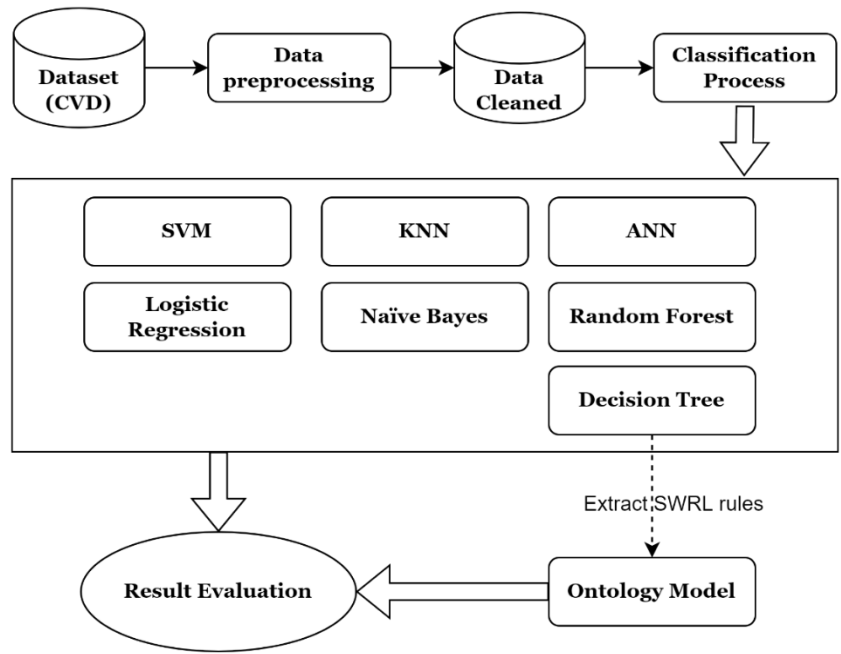

**Fig. 1.**  Experimental workflow

### 3.1 Data preprocessing

The dataset used is cardiovascular disease from Kaggle website [26], it consists of 70000 instances and 12 features (11 attributes and the last one is a target). Table 1 gives a detailed explanation of all features of the dataset.

**Table 1.** Dataset feature's information

| Attribute | Description |
|---|---|
| 1- age | The patient age (days) |
| 2- height | The patient Height (cm) |
| 3- weight | The patient Weight (Kg) |
| 4- gender | The patient Gender (Male or Female) |
| 5- ap_hi | Systolic blood pressure |
| 6- ap_lo | Diastolic blood pressure |
| 7- cholesterol | The patient Cholesterol (1: normal, 2: above normal, 3: well above normal) |
| 8- gluc | Glucose (1: normal, 2: above normal, 3: well above normal) |
| 9- smoke | The patient is smoking or not (binary) |
| 10- alco | The patient is taking alcohol or not (binary) |
| 11- active | The patient is active or not (binary) |
| 12- cardio | Target Variable (0 or 1). |

To create an effective machine learning classifier, we should always begin by data cleaning, normalizing features, transforming features, and even creating new features from the dataset. Our dataset contains 24 similar instances, after removing duplicated instances the remaining is 69976 instances. Where 35004 represents the absence of cardiovascular disease and 34972 represents the presence of the latter.

We did not add any new feature like BMI for the reason that is not given much difference in terms of results. We would like to inform you that in order to provide a fair comparison of the classification results obtained, we did not use any feature selection or performance-boosting methods.

### 3.2 Machine learning algorithms

We have used weka software for all machine learning algorithms to predict the disease. Weka contains tools for data preparation, classification, regression, clustering, association rules mining, and visualization.

We used the seven most classifiers used to classify binary datasets (Decision Tree, Random Forest, Logistic Regression, Artificial Neural Network, Naïve Bayes, Support Vector Machine, k-Nearest Neighbours). In addition, we used two modes of test options: 10-fold crossvalidation and percentage split (split 60% train, remainder test) for the reason of enriching the study.

### 3.3 Ontology model

The method used to classify the dataset using the ontology model was previously published and discussed in this earlier study [2], which we recommend reading for more information. We'll go through some specifics shortly here.

**Ontology construction.** We used Protégé software to build the ontology, it's an open-source platform that offers a suite of tools to a growing user community for building domain models and knowledge-based applications with ontologies [27]. We created the ontology manually; the main classes are Diagnostic and Patient. Figure 2 illustrates the graphical representation of the ontology.

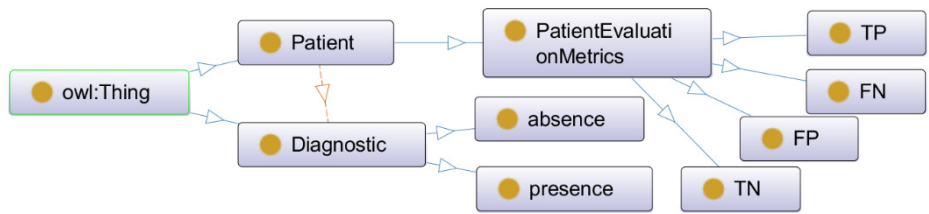

**Fig. 2.** The ontology graph

**Data properties and Instances**

The data properties used in the ontology are the same attributes presented in Table 1 which are used to build models of machine learning algorithms. Figure 3 illustrates the data properties.

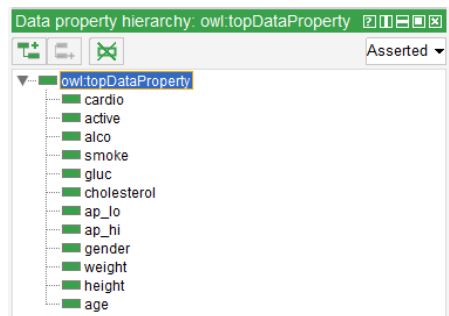

**Fig. 3.** Data properties

Cellfie, a Protégé plugin for importing spreadsheet data into OWL ontologies, is used to import the same dataset used in Weka.

**Semantic Web Language Rules (SWRL).** Following the creation of classes, data properties, and instances in the ontology. We need to establish the SWRL reasoning rules. To achieve this, we used the SWRLTab plugin, we retrieved created rules from the Decision Tree algorithm and imported them into Protégé. A java programing language is used to convert the extracted rules from the DT algorithm, where each leaf of the tree was extracted as a single SWRL rule. For instance:

```
A leaf from the DT algorithm
If cholesterol = 2 && alco ≤ 0 && smoke ≤ 0 && active ≤
0 && weight ≤ 72 && ap_lo ≤ 85 && height ≤ 169 THEN put
the patient in presence
SWRL obtained
Patient(?pt) ^ cholesterol(?pt, ?CH) ^ swrlb:equal(?CH,
'2'^^xsd:decimal) ^ alco(?pt, ?AC) ^
swrlb:lessThanOrEqual(?AC, '0'^^xsd:decimal) ^ smoke(?pt,
?S) ^ swrlb:lessThanOrEqual(?S, '0'^^xsd:decimal) ^ ac-
tive(?pt, ?A) ^ swrlb:lessThanOrEqual(?A,
'0'^^xsd:decimal) ^ weight(?pt, ?W) ^
swrlb:lessThanOrEqual(?W, '72'^^xsd:decimal) ^ ap_lo(?pt,
?AL) ^ swrlb:lessThanOrEqual(?AL, '85'^^xsd:decimal) ^
height(?pt, ?H) ^ swrlb:lessThanOrEqual(?H,
'169'^^xsd:decimal) → presence
```

**Pellet reasoned.** We utilized the Pellet reasoner, which provides more direct capabilities for working with OWL and SWRL rules, to execute SWRL rules and infer new ontology axioms. It employs the dataset and SWRL rules to instigate the inference and delivers the final decision as to whether is absence or presence of cardiovascular disease. The ontology classifier's results are reported in the next section.

### 3.4 Evaluation

Different performance metrics are used to evaluate machine learning algorithms such as Accuracy, Precision, Recall, F-Measure, ROC Area, Kappa statistic, Root mean squared error, Root relative squared error, etc. To evaluate our experimental results, we have used K-fold cross-validation and split-test with different metrics like Accuracy, Precision, Recall, and F-Measure which are described below.

Accuracy (ACC): is computed as the number of all correct predictions divided by the total number of the dataset, which is the number of patients that are identified correctly in total in our case.

$$ACC = \frac{TP+TN}{TP+TN+FP+FN} \qquad (1)$$

Precision (PREC): is calculated by dividing the number of correct positive predictions by the total number of positive predictions.

$$PREC = \frac{TP}{TP+FP} \qquad (2)$$

Recall (REC): is computed as the number of correct positive predictions divided by the total number of positives. It represents the relevant patients that have been correctly detected, it is also called Sensitivity or true positive rate (TPR).

$$REC = \frac{TP}{TP+FN} \qquad (3)$$

F-Measure: called also F-score, is a harmonic mean of precision and recall, it provides the quality of prediction.

$$\text{F-Measure} = 2 * \frac{PREC*REC}{PREC+REC} \quad (4)$$

Other metrics, such as Mean Squared Error (MSE), Root Mean Squared Error (RMSE), and Mean Absolute Error (MAE), are available but are most commonly employed in regression issues. As a result, due to the dataset and algorithms used in classification issues, this comparison research will rely on the performance metrics described above. Furthermore, the same criteria are utilized to assess the validity of our ontology model.

## 4 Results and discussion

We present in this section the results of the evaluation of classifiers employed in this study, including the result and statistics of the ontology classifier. The results of the ontology classifier are provided in Tables 2, 3 and Figure 4 using the performance metrics described in the preceding section. In addition, we provide the results of Accuracy, Precision, Recall, and F-Measure in Figures 5, 6, 7, and 8, which illustrate the visual of each metric. Table 4 further outlines the experimental results for the ML and ontology classifiers that were employed in this research.

Table 2. Ontology classifier based on 10-fold cross-validation mode

| Confusion matrix | | Actual class | |
|---|---|---|---|
| | | *positive* | *negative* |
| Predicted class | positive<br>negative | TP : 35525<br>FN : 7660 | FP : 9502<br>TN : 17289 |

Table 3. Ontology classifier based on 60% split mode validation

| Confusion matrix | | Actual class | |
|---|---|---|---|
| | | *positive* | *negative* |
| Predicted class | positive<br>negative | TP : 35681<br>FN: 7720 | FP : 9295<br>TN: 17280 |

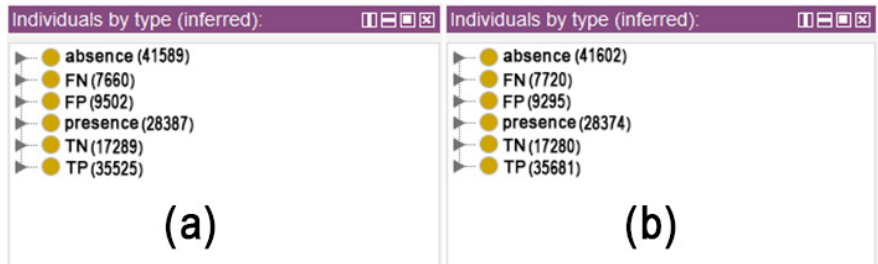

**Fig. 4.** Results of inferred concepts. (a) 10-fold cross-validation. (b) 60% split mode validation

**Accuracy.** In Figure 5 and Table 4, we obtained the highest value in terms of 10-fold cross-validation mode for Ontology, Decision Tree and Logistic Regression with 75.5%, 73.1%, 72.1% correspondingly. Almost the same results using split test mode, we obtained 75.7%, 73.1%, and 72.3% for Ontology, Decision Tree, and Logistic Regression consecutively.

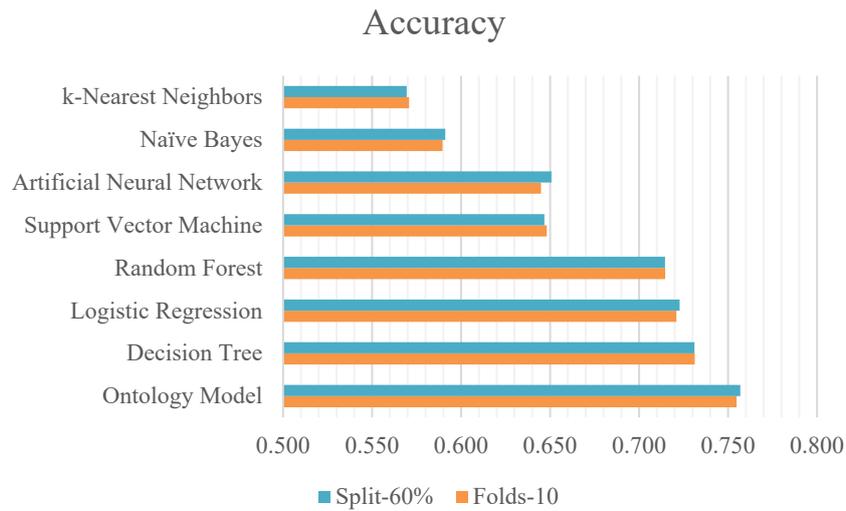

**Fig. 5.** Comparison results of accuracy

**Precision.** The ontology classifier has the highest Precision of 78.9% and 79.3% for both test modes. Followed by DT and RF. More details are shown in Table 4 and Figure 6.

*Paper*—The Impact of Ontology on the Prediction of Cardiovascular Disease Compared to Machine…

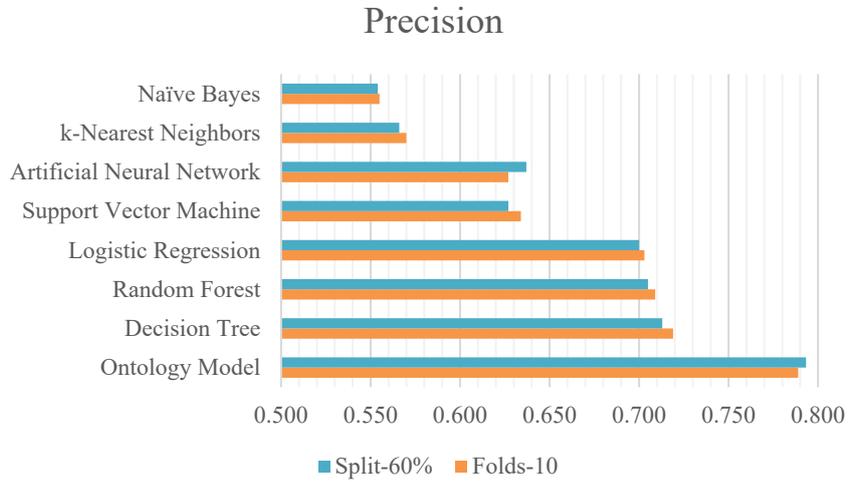

**Fig. 6.** Comparison results of precision

**Recall.** From Figure 7 and Table 4, we notice that Naïve Bayes had the highest value in both test modes, followed by Ontology in the second position and Logistic Regression with Decision Tree in the third position.

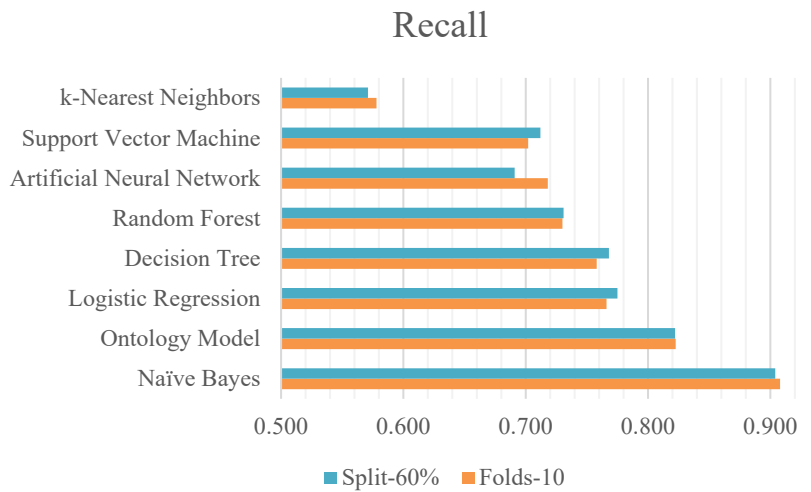

**Fig. 7.** Comparison results of recall

**F-Measure.** From Figure 8 and Table 4, we notice that the ontology model had the highest value in both test modes, followed by Decision Tree with Logistic Regression in the second position and Random Forest in the third position.

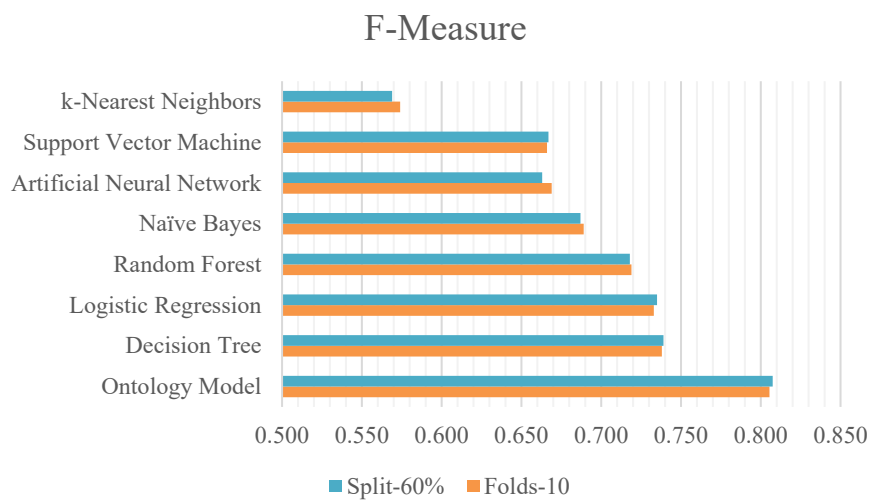

**Fig. 8.** Comparison results of F-Measure

**Table 4.** Machine learning and ontology classifiers results

|  | Accuracy | | Precision | | Recall | | F-Measure | |
| --- | --- | --- | --- | --- | --- | --- | --- | --- |
|  | *Folds-10* | *Split-60%* | *Folds-10* | *Split-60%* | *Folds-10* | *Split-60%* | *Folds-10* | *Split-60%* |
| KNN | 0,571 | 0,569 | 0,57 | 0,566 | 0,578 | 0,571 | 0,574 | 0,569 |
| NB | 0,590 | 0,591 | 0,555 | 0,554 | 0,908 | 0,904 | 0,689 | 0,687 |
| ANN | 0,645 | 0,651 | 0,627 | 0,637 | 0,718 | 0,691 | 0,669 | 0,663 |
| SVM | 0,648 | 0,647 | 0,634 | 0,627 | 0,702 | 0,712 | 0,666 | 0,667 |
| RF | 0,715 | 0,715 | 0,709 | 0,705 | 0,73 | 0,731 | 0,719 | 0,718 |
| LR | 0,721 | 0,723 | 0,703 | 0,7 | 0,766 | 0,775 | 0,733 | 0,735 |
| DT | 0,731 | 0,731 | 0,719 | 0,713 | 0,758 | 0,768 | 0,738 | 0,739 |
| Ontology | 0,755 | 0,757 | 0,789 | 0,793 | 0,823 | 0,822 | 0,805 | 0,807 |

Regarding results discussed above, we notice that there is no big difference between cross-validation and percentage split test mode. The experimental results show that the ontology classifier is considered the best with high accuracy of 75.5%, followed by the Decision Tree of 73.1% and logistic regression of 72.1%. We conclude that combining machine learning with ontological reasoning (i.e., extracting rules from machine learning algorithms and integrating them into the ontology using SWRL) may provide better outcomes. Furthermore, these comparative findings demonstrate how OWL ontology's knowledge representation and reasoning capabilities might bring further benefits in addition to classification. Furthermore, because the ontology classifier is an interpretable model, it may offer information on how the process reaches the decision. The ontology classifier produces equivalent and comparable results to machine learning classifiers. The findings may also be interpreted by humans, and the rules can be altered or added as needed.

To our knowledge, this is the first comparative analysis of ML and ontology classifiers, in which we have integrated ontology with machine learning and specifically in the field of prediction of cardiovascular disease. So, no meaningful comparison can be made for that reason on the one hand, on the other hand, researchers use different datasets and different features selection and performance-boosting methods.

## 5    Conclusion

Machine learning techniques are widely employed in all scientific disciplines and have revolutionized industries all over the world. The application of machine learning tools and algorithms in healthcare has lately witnessed significant advancement [28-30]. Those methods have demonstrated efficacy and may be beneficial in the treatment of chronic diseases such as cardiovascular disease. In addition, the Semantic Web, for its part, has demonstrated its worth and strength in a variety of disciplines, including health. Ontology, as a component of the Semantic Web, comes with the capacity to process concepts and relationships in the same manner that humans see connected concepts.

In this paper, we presented seven machine learning algorithms and an ontology model, and we explained their comparative evaluation. In addition, different performance metrics are used to evaluate the results such as Accuracy, Precision, Recall, F-Measure.

The findings reveal that, even with no feature selection applied, the ontology classification method has the highest accuracy. This leads us to a new search field that we suggest and encourage researchers to contribute and create new ideas in the same context, to give more results and comparison, for the purpose of prediction, recommendation, or make a decision, etc. From our end, we look forward to improving this comparative study by applying new approaches to integrate rules of machine learning with the ontology classification method, as well as using regression machine learning algorithms.

## 7    Authors

**Noreddine Gherab** is a professor of computer science with industrial and academic experience. He holds a doctorate degree in computer science. In 2013, he worked as a professor of computer science at Mohamed Ben Abdellah University and since 2015 has worked as a research professor at Sultan Moulay Slimane University, Morocco. Member of the International Association of Engineers (IAENG). Professor Gherabi having several contributions in information systems namely: big data, semantic web, pattern recognition, intelligent systems. He has papers (book chapters, international journals, and conferences/workshops), and edited books. He has served on executive and technical program committees and as a reviewer of numerous international conferences and journals, he convened and chaired more than 30 conferences and workshops. He is member of the editorial board of several other renowned international journals: • Co-editor in chief (Editorial Board) in the journal "The International Journal of sports science and engineering for children" (IJSSEC). • Associate Editor in the journal « International Journal of Engineering Research and Sports Science ». • Reviewer in several journals / Conferences • Excellence Award, the best innovation in science and technology 2009. His research areas include Machine Learning, Deep Learning, Big Data, Semantic Web, and Ontology. He can be contacted at email: n.gherabi@usms.ma.

**Hakim El Massari** received his master degree from Normal Superior School of Abdelmalek Essaadi University, Tétouan, Morocco, in 2014. Currently, he is preparing his Ph.D. in computer science at the National School of Applied Sciences, Sultan Moulay Slimane University, Khouribga, Morocco. His research areas include Machine Learning, Deep Learning, Big Data, Semantic Web, and Ontology. He can be contacted at email: h.elmassari@usms.ma.

**Hamza Ghandi** is pursuing her Ph.D. in Computer Engineering from the National School of Applied Sciences of Khouribga, her area of interest is Machine Learning, Intelligent Systems, Deep Learning, and Big Data.

**Sajida Mhammedi** received her Ms Degree in Computer Engineering from Faculty of Science and Technologie, Beni Mellal Morocco, she worked as a visiting researcher at the Sultane Moulay Slimane University, her research interests include Machine Learning, Semantic Web, recommendation systems, Ontology, and Big Data.

**Mohamed Bahaj** obtained his Ph.D. in Mathematics and Computer Science from University Hassan 1st, Morocco. He is co-chairs of International Conferences on International Conference on Intelligent Information and Network Technology. He is a Full Professor in Department of Mathematics and Computer Sciences from the University Hassan 1st Faculty of Sciences & Technology Settat Morocco. He has published over 120 peer-reviewed papers. His research interests are Intelligent Systems, Ontologies Engineering, Partial and differential Equations, Numerical Analysis and Scientific Computing He has reviewed several papers in various journals. He has supervised several PhD thesis in Computers Sciences & in Mathematics. He is a Guest Editorial in Journal of Emerging Technologies in Web Intelligence. He also attended a series of workshops, seminars and discussion forums for Academic Development on teaching and research.